\documentclass{WileyMSP-template}

\usepackage[colorlinks,
            linkcolor={red!100!black},
            citecolor={blue!100!black},
            urlcolor={blue!80!black}]{hyperref}
\usepackage{xurl}            % simple URL typesetting
\usepackage{booktabs}       % professional-quality tables
\usepackage{nicefrac}       % compact symbols for 1/2, etc.
\usepackage{multirow}

\usepackage[utf8]{inputenc} % allow utf-8 input
\usepackage[T1]{fontenc}    % use 8-bit T1 fonts
\usepackage{hyperref}
\usepackage[font=footnotesize,labelfont=bf ]{caption}
\usepackage{bbm}
\usepackage{bm}
\usepackage{amsmath,amssymb,amsfonts,amsmath}
\usepackage{cleveref}
\usepackage{wrapfig}
\usepackage{float}
\usepackage{dsfont}
\usepackage{graphicx}
\usepackage{microtype}
\usepackage{setspace}
\usepackage{soul}
\usepackage{tikz}
\usepackage{amsmath,amsthm,amssymb,amsfonts}
\usepackage{upgreek}
\usepackage{textgreek}
\usepackage{mathtools,mathdots}
\usepackage{xparse}
\usepackage{nameref}
\usepackage{graphicx}
\usepackage{subcaption}
\usepackage{float}
\usepackage{nicefrac}
\usepackage{multirow}
\usepackage{multicol}
\usepackage{hyperref}
\usepackage{multimedia}
\usepackage{arydshln}
\usepackage{substr}
\usepackage{mathrsfs}
\let\oldmathcal=\mathcal
\renewcommand{\mathcal}[1]{
    \IfSubStringInString{#1}{ABCDEFGHIJKLMNOPQRSTUVWXYZ}{\oldmathcal{#1}}{
    \IfSubStringInString{#1}{abcdefghijklmnopqrstuvwxyz}{\mathsf{#1}}{
    \ifthenelse{\equal{#1}{\epsilon}}{\textnormal{\straightepsilon}}{
    #1
    }}}
}

\usepackage{algorithm}
\usepackage{algpseudocode}

\algrenewcommand\alglinenumber[1]{#1:}

\usepackage{pbox}

\DeclareMathOperator*{\minimize}{minimize}

\newcommand{\Mat}[1][]{\ifthenelse{\equal{#1}{}}{\text{Mat}}{\text{Mat}(#1)}}

\newcommand{\SE}[1]{\text{SE}(#1)}

\newcommand{\TSE}[2][]{
	\ifthenelse{\equal{#1}{}}
	{{T\SE{#2}}}
	{{T_{#1}\SE{#2}}}
}
\newcommand{\dualTSE}[2][]{
	\ifthenelse{\equal{#1}{}}
	{{T^*\SE{#2}}}
	{{T^*_{#1}\SE{#2}}}
}

\newcommand{\dif}{\mathrm{d}}

\newcommand{\abs}[1]{\left|#1\right|}

\newcommand{\Vector}[1]{#1}

\usepackage{xcolor}
\usepackage{algorithm}
\usepackage{algpseudocode}

\algrenewcommand\alglinenumber[1]{#1:}

\newcommand{\positions}{\Vector{r}}

\newcommand{\strain}{\epsilon}

\newcommand{\shears}{\Vector{\nu}}

\newcommand{\internalforce}{n}
\newcommand{\internalforces}{\Vector{n}}

\newcommand{\internalcouples}{\Vector{m}}

\newcommand{\muscle}{\text{m}}

\newcommand{\musclepositions}[1][]{\positions^{\ifthenelse{\equal{#1}{}}{\muscle}{#1}}}
\newcommand{\musclerelativepositions}[1][]{\Vector{\gamma}^{\ifthenelse{\equal{#1}{}}{\muscle}{#1}}}

\newcommand{\musclelength}[1][]{\ell^{\ifthenelse{\equal{#1}{}}{\muscle}{#1}}}
\newcommand{\musclestrain}[1][]{\strain^{\ifthenelse{\equal{#1}{}}{\muscle}{#1}}}
\newcommand{\muscleshears}[1][]{\shears^{\ifthenelse{\equal{#1}{}}{\muscle}{#1}}}
\newcommand{\muscletangent}[1][]{\Vector{t}^{\ifthenelse{\equal{#1}{}}{\muscle}{#1}}}
\newcommand{\muscleforce}[1][]{\internalforce^{\ifthenelse{\equal{#1}{}}{\muscle}{#1}}}
\newcommand{\maxmusclestress}[1][]{\sigma^{\ifthenelse{\equal{#1}{}}{\muscle}{#1}}}
\newcommand{\maxmuscleforce}[1][]{\internalforce^{\ifthenelse{\equal{#1}{}}{\muscle}{#1}}_\text{max}}
\newcommand{\muscleforces}[1][]{\internalforces^{\ifthenelse{\equal{#1}{}}{\muscle}{#1}}}
\newcommand{\musclecouples}[1][]{\internalcouples^{\ifthenelse{\equal{#1}{}}{\muscle}{#1}}}
\newcommand{\muscleactivation}[1][]{a^{\ifthenelse{\equal{#1}{}}{\muscle}{#1}}}
\newcommand{\staticmuscleactivation}[1][]{\alpha^{\ifthenelse{\equal{#1}{}}{\muscle}{#1}}}

\newcommand{\musclestoredenergy}[1][]{W^{\ifthenelse{\equal{#1}{}}{\muscle}{#1}}}

\newcommand{\LM}[1][]{\text{LM}{\ifthenelse{\equal{#1}{}}{}{_{#1}}}}
\newcommand{\OM}[1][]{\text{OM}{\ifthenelse{\equal{#1}{}}{}{_{#1}}}}

\def\R{{\mathds{R}}}

\def\0{{\mathbb{0}}}
\def\1{{\mathds{1}}}

\def\a{{\mathbf{a}}}
\def\b{{\mathbf{b}}}

\definecolor{db}{RGB}{23,20,119}
\definecolor{dg}{RGB}{2,101,15}

\usepackage{upgreek}

\newcommand{\material}[1]{
	\ifthenelse{\equal{#1}{\kappa}}{\upkappa}{
	\ifthenelse{\equal{#1}{\nu}}{\upnu}{
	\ifthenelse{\equal{#1}{\omega}}{\upomega}{
	\ifthenelse{\equal{#1}{\sigma}}{\upsigma}{
	\ifthenelse{\equal{#1}{\theta}}{\uptheta}{
	\mathsf{#1}}}}}}
}

\renewcommand{\a}{\mathsf{a}}
\renewcommand{\b}{\mathsf{b}}

\usepackage{sectsty}
\subsubsectionfont{\fontsize{10}{15}\selectfont}
\sectionfont{\LARGE}
\subsectionfont{\Large}

\usepackage[indent=10pt]{parskip}
\usepackage[fontsize=9pt]{fontsize}

\begin{document}
\pagestyle{fancy}
\title{Hierarchical control and learning of a foraging CyberOctopus}
\maketitle

% Author: Please give full first and last names for authors and include * after the name of all corresponding authors
\author{Chia-Hsien Shih,} 
\author{Noel Naughton,}
\author{Udit Halder,}
\author{Heng-Sheng Chang,} \\[4pt]
\author{Seung Hyun Kim,}
\author{Rhanor Gillette,}
\author{Prashant G. Mehta,}
\author{Mattia Gazzola\textsuperscript{*}}

% Affiliations: Please provide academic titles (Prof. or Dr.) for all authors where applicable, and include an institutional email address for all corresponding authors
\begin{affiliations}
Chia-Hsien Shih, Heng-Sheng Chang, Seung Hyun Kim, Prashant G. Mehta, and Mattia Gazzola\\
Department of Mechanical Science and Engineering \\ University of Illinois at Urbana-Champaign\\
Urbana, IL 61801\\
Email Address: mgazzola@illinois.edu

Noel Naughton\\
Beckman Institute for Advanced Science and Technology\\
University of Illinois at Urbana-Champaign\\
Urbana, IL 61801

Udit Halder\\
Coordinated Science Laboratory\\
University of Illinois at Urbana-Champaign\\
Urbana, IL 61801

Rhanor Gillette\\
Department of Molecular and Integrative Physiology\\
University of Illinois at Urbana-Champaign\\
Urbana, IL 61801

\end{affiliations}

% Keywords: Please provide a minimum of three and a maximum of seven keywords, separated by commas

% \linespacing
\justifying
% Abstract should be written in the present tense and impersonal style (i.e., avoid we), and be at most 200 words long
\begin{abstract}
    % \linespacing
    
\normalsize{\textbf{Abstract.} Inspired by the unique neurophysiology of the octopus, we propose a hierarchical framework that simplifies the coordination of multiple soft arms by decomposing control into high-level decision making, low-level motor activation, and local reflexive behaviors via sensory feedback. When evaluated in the illustrative problem of a model octopus foraging for food, this hierarchical decomposition results in significant improvements relative to end-to-end methods. Performance is achieved through a mixed-modes approach, whereby qualitatively different tasks are addressed via complementary control schemes. Here, model-free reinforcement learning is employed for high-level decision-making, while model-based energy shaping takes care of arm-level motor execution. To render the pairing computationally tenable, a novel neural-network energy shaping (NN-ES) controller is developed, achieving accurate motions with time-to-solutions 200 times faster than previous attempts. Our hierarchical framework is then successfully deployed in increasingly challenging foraging scenarios, including an arena littered with obstacles in 3D space, demonstrating the viability of our approach.}

\end{abstract}

%%%%%%%%%%%%%%% End of first page %%%%%%%%%%%%%%%%%%%%%

% \author{
%     {Chia-Hsien Shih\textsuperscript{1},} 
%     {Noel Naughton\textsuperscript{2},}
%     {Udit Halder\textsuperscript{3},}
%     {Heng-Sheng Chang\textsuperscript{1,3},} \\[4pt]
%     {Seung Hyun Kim\textsuperscript{1},}
%     {Rhanor Gillette\textsuperscript{4},}
%     {Prashant G. Mehta\textsuperscript{1,3},}
%     {Mattia Gazzola\textsuperscript{1,5,6,*}}
% }
% \date{}
% \maketitle
% % \clearpage

% %%%%%%%%% Insert author address here
% % \address{
% \vspace{-15pt}
% \small
% \noindent
% \textsuperscript{1}Mechanical Science and Engineering, University of Illinois at Urbana-Champaign, \\[4pt]
% \textsuperscript{2}Beckman Institute for Advanced Science and Technology, University of Illinois at Urbana-Champaign, \\[4pt]
% \textsuperscript{3}Coordinated Science Laboratory, University of Illinois at Urbana-Champaign, \\[4pt]
% \textsuperscript{4}Department of Molecular and Integrative Physiology, University of Illinois at Urbana-Champaign, \\[4pt]
% \textsuperscript{5}National Center for Supercomputing Applications, University of Illinois at Urbana-Champaign, \\[4pt]
% \textsuperscript{6}Carl R. Woese Institute for Genomic Biology, University of Illinois at Urbana-Champaign\\[4pt]
% \textsuperscript{*}Correspondence: Mattia Gazzola (mgazzola@illinois.edu) \\

\section{Introduction}

Soft materials have long been seen as key drivers for improving robots' abilities to interact with and adapt to their environments \cite{Sadeghi2017,Zhang:2019,Huang:2019}. Nonetheless, it is precisely the advantages afforded by physical compliance that, in turn, render soft robots difficult to control. Indeed, material non-linearities, instabilities, continuum mechanics, distributed actuation, and conformability to the environment all render the control problem challenging \cite{DellaSantina2021,GeorgeThuruthel2018}. In search of potential solutions, roboticists have turned to nature for inspiration \cite{Trivedi:2008}, from the investigation of elephant trunks \cite{Hannan:2001,Baumgartner:2020}, mammalian tongues \cite{Noel:2018,Lu:2015}, and inchworms \cite{Joyee:2019,Gamus:2020,Karipoth:2022} to plant tendrils \cite{Must:2019,Skotheim:2005}, starfish \cite{Otake:2002, Jin:2016}, snakes \cite{Zhang:2021,Marvi:2014}, and octopuses \cite{Li2012a,Cianchetti2015,Grissom2006,Wu:2021,Sakuhara:2020,Chang:2021,Walker:2005}. The latter, in particular, have received a great deal of attention owing to their unmatched ability to orchestrate multiple soft arms for swimming, crawling, and hunting, as well as manipulating complex objects, from coconut shells to lidded jars \cite{Grasso:2008,Kennedy:2020}. 

Within this context, a potential solution framework based on hierarchical decomposition is suggested by the unique neurophysiology of the octopus. In contrast to the mostly centralized brain structure of vertebrates \cite{Godfrey:2017}, the octopus exhibits a highly distributed neural system wherein two thirds of its brain lies within its arms \cite{Young:1971}. This Peripheral Nervous System (PNS) is organized into brachial ganglia, colocated with the suckers, and is responsible for low-level sensorimotor tasks and whole-arm motion coordination \cite{Hochner:2012}. Indeed, surgically severed arms are known for being able to execute motor programs such as reaching or recoiling \cite{Sumbre:2001, Hague:2013}. 
The Central Nervous System (CNS), composed of the remaining third of the neural tissue, is instead located in the mantle and is thought to be responsible for learning and decision making by integrating signals from the entire body \cite{Gutnick2020}. This neural architecture is naturally suggestive of a control hierarchy wherein high-level decisions are made in the CNS, executed by the PNS, and finally modulated by the local environment via arm compliance. 

\begin{figure*}[t!]
\centering
\includegraphics[width=1.0\textwidth]{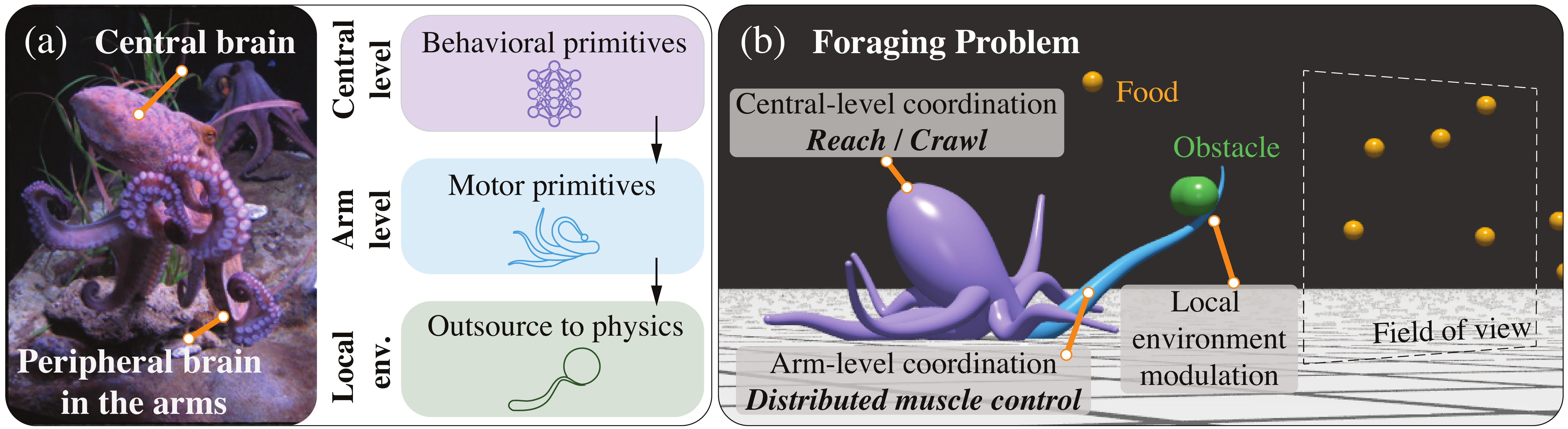}
\captionsetup{belowskip=-10pt}
\caption{(a) Our proposed control hierarchy inspired by the distributed neurophysiology of the octopus: a centralized decision-maker selects appropriate motion primitives, arm-level controllers generate the necessary muscle activations and physical compliance accommodates for environmental obstacles. (b) A CyberOctopus foraging for food in the presence of obstacles. }
\label{fig:problem}
\end{figure*}

Reflecting these considerations, we present a tri-level hierarchical approach to coordination and learning in an octopus computational analog, henceforth referred to as the CyberOctopus. Our framework, illustrated in Fig.~\ref{fig:problem}, decomposes control into central-level, arm-level, and local environment-level.
At the central level, executive and coordination decisions (behavioral primitives), such as reaching for food or crawling, are taken and issued to the individual arms. This top level is implemented as a compact, feedforward neural network. 
At the level of the arm, modeled as an elastic slender filament, muscle activations (motor primitives) realize incoming commands and produce appropriate deformations \cite{Chang:2021}. Muscle control is obtained via a fast energy shaping technique that minimizes energy expenditure \cite{Chang:2020_ES}. 
We supplement this control with distributed, local behavioral rules that conspire with the arm's compliant physics to autonomously accommodate for solid environmental features.

The CyberOctopus is shown to learn to forage for food in increasingly challenging scenarios, including an arena littered with obstacles in 3D space. Overall, this work illustrates how hierarchical control is not only viable in soft multi-arms systems but, in fact, can significantly outperform end-to-end, deep-learning approaches.

\section{The CyberOctopus model}\label{sec:cyberoctopus_model}

\begin{figure*}[t!]
\centering
\includegraphics[width=1.0\textwidth]{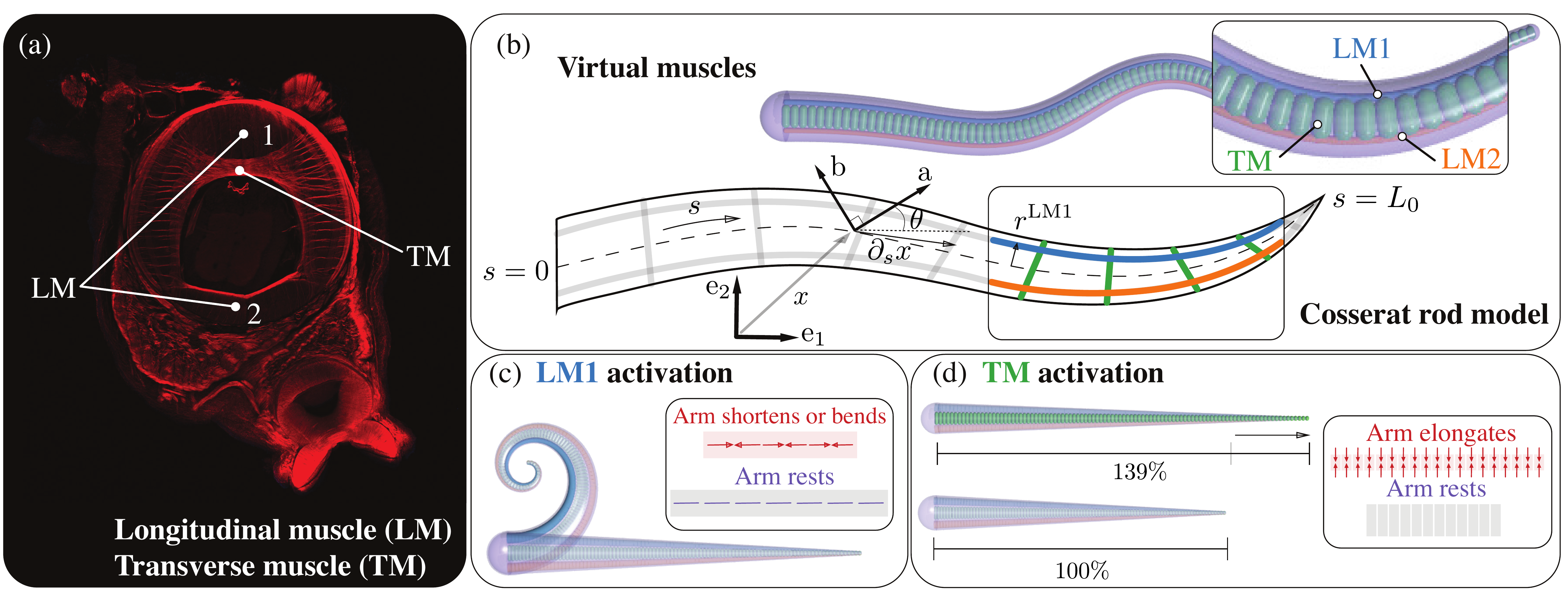}
\captionsetup{belowskip=-10pt}
\caption{(a) Histological cross-section of an \textit{Octopus rubescens} arm showing the longitudinal (LM1 and LM2) and transverse (TM) muscles. Muscles are labeled in red by phalloidin staining (Image credit: Tigran Norekian and Ekaterina D. Gribkova).
(b) Accordingly, our model arm consists of top (LM1, blue) and bottom (LM2, orange) virtual longitudinal muscles as well as virtual transverse muscles (TM, green). The soft arm itself is represented as a single Cosserat rod, to capture its passive elastic mechanics. 
Muscle activations are defined along the arm. (c) Longitudinal muscle activations result in arm shortening (symmetric co-contraction) or bending (asymmetric co-contraction), while (d) transverse muscle activations result in arm elongations, due to tissue incompressibility.
}
\label{fig:model}
\end{figure*}

Of all the elements that comprise a real octopus, here we focus on its arms and their coordination. 
The arms of the CyberOctopus (Fig.~\ref{fig:model}) are modeled as linearly tapered Cosserat rods, which are slender, one-dimensional elastic structures that can undergo all modes of deformation --- stretch, bend, twist, and shear --- at every cross section \cite{Gazzola2018a}. Each arm is then represented as an individual passive rod upon which virtual muscles produce forces and couples. We consider each arm deforming in-plane on account of two longitudinal muscle groups (LM1 and LM2) and one set of transverse muscles (TM), reflecting the octopus' physiology, as illustrated in Fig.~\ref{fig:model}a,b. 
% that can each be independently contracted along the length of the arm $s$.
Longitudinal muscles (Fig.~\ref{fig:model}c) are located off-center from the arm's axis of symmetry but run parallel to it, generating both forces and couples that can cause the arm to contract (symmetric co-contractions) or bend (asymmetric co-contractions). Transverse muscles (Fig.~\ref{fig:model}d) are located along the arm's axis but are oriented orthogonally, so that their contraction causes the arm to extend due to incompressibility. 
Oblique muscles, whose main function is to provide twist \cite{Kier:2007}, are not considered here as they are not relevant for planar motion.

\textbf{Kinematics.} In the Cosserat rod formalism (Fig.~\ref{fig:model}b), each arm is described by its midline position vector $x(s, t) \in \R^2$ within the plane spanned by the fixed orthonormal basis $\{\mathsf{e}_1, \mathsf{e}_2\}$ and along the arclength $s \in [0,L_0]$, where $L_0$ is the arm's rest length, and $t$ is time. The arm's local orientation is described by the angle $\theta(s,t) \in \R$ which defines the local orthonormal basis $\{\a, \b\}$ with $\a = \cos \theta \,\mathsf{e}_1 + \sin \theta \, \mathsf{e}_2$ and $\b = -\sin \theta \, \mathsf{e}_1 + \cos \theta \, \mathsf{e}_2$. The local deformations of the arm --- stretch ($\nu_1$), shear ($\nu_2$), and bending ($\kappa$) --- are defined by the kinematics of the arm
\begin{align}
	\begin{split}
		\partial_s x = \nu_1 \a +\nu_2 \b, \quad\quad  
		\partial_s \theta = \kappa 
% 		\frac{\partial x}{\partial s}  &= \nu_1 \a +\nu_2 \b \\ 
% 		\frac{\partial \theta}{\partial s}  &= \kappa
			\label{eq:state_kinematics}
	\end{split}
\end{align}
For a straight arm at rest, $\nu_1=1$ and $\nu_2=\kappa=0$. 

\textbf{Dynamics.} The dynamics of the planar arm \cite{antman1995nonlinear, Chang:2021} read  
\begin{align}
	\begin{split}
		     \partial_t \begin{bmatrix}
             \rho A v_1\\
             \rho A v_2\\
             \rho I \omega\\
            \end{bmatrix}
            =
		    \begin{bmatrix} \partial_s \left( \begin{bmatrix*}[r] \cos \theta & -\sin \theta \\
		                                                        \sin \theta & \cos \theta \end{bmatrix*}
            \begin{bmatrix*} n_1 \\
            n_2 \end{bmatrix*} \right) \\ 
            \partial_s m + \nu_1 n_2 - \nu_2 n_1 \\
            \end{bmatrix} 
             - \zeta 
		    \begin{bmatrix}
             v_1\\
             v_2\\
           \omega \\
            \end{bmatrix}
	\end{split}
	\label{eq:CR_dynamics}
\end{align}
where $v=(v_1,v_2)$ and $\omega$ are the linear and angular velocity of the arm, respectively, $\rho$ is the density of the arm, $A$ and $I$ are the arm's local cross-sectional area and second moment of area, $n=(n_1,n_2)$ and $m$ are the internal forces (in the local frame) and couples along the arm, and $\zeta >0$ is a damping coefficient capturing viscoelastic effects. 
The arm dynamics (Eq.~\ref{eq:CR_dynamics}) are accompanied by a set of fixed/free type boundary conditions for all $t\geq0$
\begin{align}
\begin{split}
\text{(fixed)} \qquad &x(0,t) = x_0, \ \ \theta(0,t) = \theta_0 \\
\text{(free)} \qquad   &n(L_0, t) = 0, \ \ m(L_0, t) = 0
\end{split}
\end{align}
where $x_0 \in \R^2$ and $\theta_0 \in \R$ are the prescribed position and orientation of the arm at its base. Since the arm is freely moving, a free boundary condition at the tip is chosen.

\textbf{Internal stresses.}
The overall internal forces and couples acting on the arm ($n$, $m$) encompass both passive and active effects
\begin{align}
n = n^{\text{e}} + \sum_{\muscle \in M} n^\muscle, \quad m = m^{\text{e}} + \sum_{\muscle \in M} m^\muscle
\end{align}
where $(n^{\text{e}}, m^{\text{e}})$ are restoring loads due to passive elasticity, and $(n^\muscle, m^\muscle)$ are the active loads resulting from the contraction of muscle $\muscle \in M$, with $M = \{\text{LM1, LM2, TM}\}$ being the collection of muscle groups in the arm.

For a linearly elastic arm, passive elastic forces and couples read
\begin{align}
		n^{\text{e}} = \begin{bmatrix}
            EA(\nu_1-1) \\ 
            GA\nu_2
            \end{bmatrix}, \quad 
    m^{\text{e}} = EI\kappa
\end{align}
where $E$ and $G$ are Young’s and shear moduli. 

The contraction of a muscle $\muscle \in \{\text{LM1, LM2, TM}\}$ is modeled via the activation function $\alpha^\muscle(s,t) \in [0,1]$, with 1 corresponding to maximum activation (Fig.~\ref{fig:model}c,d). 
When virtual muscles contract, they produce on the arm a distributed force $n^\text{m}(s,t)$ (in the local frame) that depends on the musculature's spatial organization. We model this as
\begin{align}
	\begin{split}
		n^\muscle = \begin{bmatrix} \alpha^\muscle ~\sigma^\muscle  A^\muscle ~\mathrm{T}^\muscle \\ 0 \end{bmatrix}
		%n^\muscle =  \alpha^\muscle ~\sigma^\muscle  A^\muscle ~\mathcal{t}^\muscle
	\end{split}
	\label{eq:muscle_forces}
\end{align}
where $\sigma^\muscle$ is the maximum stress generated by the muscle, $A^\muscle$ is its cross sectional area, and $\mathrm{T}^\muscle$ accounts for the musculature configuration.
For longitudinal muscles, $\mathrm{T}^{\text{LM}}=1$ since contractions directly translate into compression forces along the arm (note that due to small shear $\nu_2\approx 0$ --- confirmed numerically --- the vector $\a$ of Fig.~\ref{fig:model}b effectively coincides with $\partial_s x$).
For transverse muscles, $\mathrm{T}^{\text{TM}}=-1$, capturing the fact that their contractions cause radial shortening, which in turn extends the arm due to incompressibility (Fig.~\ref{fig:model}d).
Additionally, the muscles' force-length relationships are modeled here as a constant function, however, more complex descriptions can be incorporated \cite{Chang:2021}. 

Due to the offset location of the longitudinal muscles with respect to the arm's main axis, the active forces $n^{\text{LM}}$ generate couples ($m^{\text{LM}}$), which are modeled as follows.
We denote the position of a muscle relative to the arm centerline by the vector $x^\muscle(s) = \pm \phi^\muscle(s) \mathsf{b}$, where $\phi^\muscle (s)$ is the off-center distance (Fig.~\ref{fig:model}b). 
The positive and negative signs are associated with LM1 and LM2, respectively.
Then the resulting couples are 
\begin{equation}
		m^\muscle = (x^\muscle \times n^\text{m})\cdot (\mathsf{e}_1 \times \mathsf{e}_2) = \pm \phi^\muscle \alpha^\muscle ~\sigma^\muscle  A^\muscle ~\mathrm{T}^\muscle
\end{equation}
Transverse muscles are arranged perpendicularly to the arm (Fig.~\ref{fig:model}d), and thus result in no couples ($m^{\text{TM}} = 0$).

\textbf{Static configurations.} For a static muscle activation $\alpha^\muscle (s)$, the equilibrium configuration of the arm is characterized by the balance of forces and couples. This is obtained by equating the right-hand side of the dynamics \eqref{eq:CR_dynamics} to zero, yielding
\begin{align}
	\begin{split}
	\underbrace{
		    \begin{bmatrix}
            n^\text{e}\\
            m^\text{e}
            \end{bmatrix}
            }_{\substack{\text{Elastic arm}\\ \text{passive response}}}
            = - 
            \underbrace{
             \sum_{\muscle\in M}
            \begin{bmatrix}
            n^\muscle\\
            m^\muscle
            \end{bmatrix}
            }_{\substack{\text{Muscle}\\ \text{contractions}}}
            % }_{\text{Muscle contractions}},
	\end{split}
	\label{eq:matchingcondition}
\end{align}
Equation \ref{eq:matchingcondition} is solved for the static strains $\nu_1, \nu_2, \kappa$, which in turn lead to the equilibrium configuration of the arm, obtained by integrating the kinematics of Eq.~\ref{eq:state_kinematics}.

While Eq.~\ref{eq:matchingcondition} suffices to determine the equilibrium configuration for given muscle activations $\alpha^\muscle(s)$, it does not account for the dynamical response of the arm transitioning between activations, all the while experiencing environmental loads.
To remedy this, we evaluate the effect of muscle activations (and thus of the control policies that determine them) in \textit{Elastica} \cite{Gazzola2018a,Zhang2019,Naughton:2021}, an open-source software for simulating the dynamics of Cosserat rods (Eq.~\ref{eq:CR_dynamics}). \textit{Elastica} has been demonstrated across a range of biophysical applications from soft \cite{Naughton:2021} and biohybrid \cite{Zhang2019,Aydin:2019,Pagan-Diaz:2018,Wang:2021} robots to artificial muscles \cite{Charles:2019} and biolocomotion \cite{Zhang2019,Zhang:2021,Chang:2020_ES,Chang:2021}. 
In \textit{Elastica}, our CyberOctopus consists of a head and eight arms, of which only a subset (gradually increased throughout the paper) is actively engaged.
Material and geometric properties of our model octopus are determined from typical literature values \cite{Tramacere:2014} as well as experimental characterizations of \textit{Octopus rubescens} \cite{Chang:2020_ES}.  
Numerical values and details of our muscle models are provided in SI Tables 1-3 and in \cite{Chang:2021}.

\section{Arm-level problem: motor execution}\label{sec:Primitives}

Octopuses perform certain goal-directed arm motions via templates of muscle activations, such as traveling waves of muscle contractions \cite{Sumbre:2001}.
These templates are encoded into the arm's peripheral nervous system as low-level motor programs that are selected, modulated, and combined together to achieve basic behaviors such as reaching and fetching \cite{Gutfreund:1998,Sumbre:2001,Sumbre:2006}.
Inspired by this, we define two types of primitives for inclusion in our hierarchical approach: \textit{motor primitives} (Sec. \ref{sec:motor_primitives}) and \textit{behavioral primitives} (Sec. \ref{sec:behavioral_primitives}). 
Motor primitives are low-level motor programs that coordinate the contraction of the CyberOctopus' muscles to accomplish a stereotypical motion. 
Behavioral primitives are sequences of motor primitives whose combination enables the completion of simple goal-directed tasks (here crawling or reaching available food). These behavioral primitives can then be further composed into more complex behaviors, such as foraging.

\subsection{Motor primitives: Reaching to a point in space}\label{sec:motor_primitives}
We focus on a motor primitive that efficiently moves the tip of the arm to a specified location $q \in \R^2$. This basic motion can be used to accomplish a variety of tasks, for example reaching to a food target, fetching food to the mouth, or crawling. 

 \textbf{Energy shaping (ES).} To effect this motor primitive, we employ the energy shaping methodology \cite{Ortega:2001,Blankenstein:2002,Bloch:2000}. 
As developed in our prior work \cite{Chang:2020_ES, Chang:2021, chang2022energy}, an energy shaping control law is derived to determine the static muscle activations $\alpha =\{\alpha^\muscle\}_{\muscle \in M}$ that cause the tip of the arm to reach a target location. The equilibrium arm configuration that achieves this goal is obtained by solving an optimization problem that minimizes the tip-to-target distance $\delta (\alpha, q) = \left| q - x(L_0) \right|$ along with the muscle activation cost \cite{Chang:2021}
\begin{align}
\label{eq:ESoptimization}
\begin{array}{c}
\text{Energy} \\ \text{Shaping } \\ \text{(ES)} 
\end{array}
\left\{\rule{0cm}{0.75cm}\right. %\quad
\minimize_{\alpha(\cdot); \, \alpha^\muscle(s) \in [0,1]} \quad & J(\alpha; \alpha_0, q) = \overbrace{\int^{L_0}_0 \left| \alpha(s)-\alpha_0(s)\right|^2 \dif s}^{\text{muscle cost}} + \overbrace{\vphantom{\int_0^{L_0}}\mu_{\text{tip}}\delta(\alpha,q)}^{\text{task specific cost}} 
\end{align}
where $\alpha_0$ are the initial muscle activations, and $\mu_{\text{tip}}$ is a constant (regularization) coefficient.
The tip-to-target distance $\delta(\alpha,q)$ is computed using the kinematic constraints of Eq.~\ref{eq:state_kinematics} and equilibrium constraints of Eq.~\ref{eq:matchingcondition}.

\begin{figure*}[t!]
    \centering
    \centering
    \includegraphics[width=\textwidth]{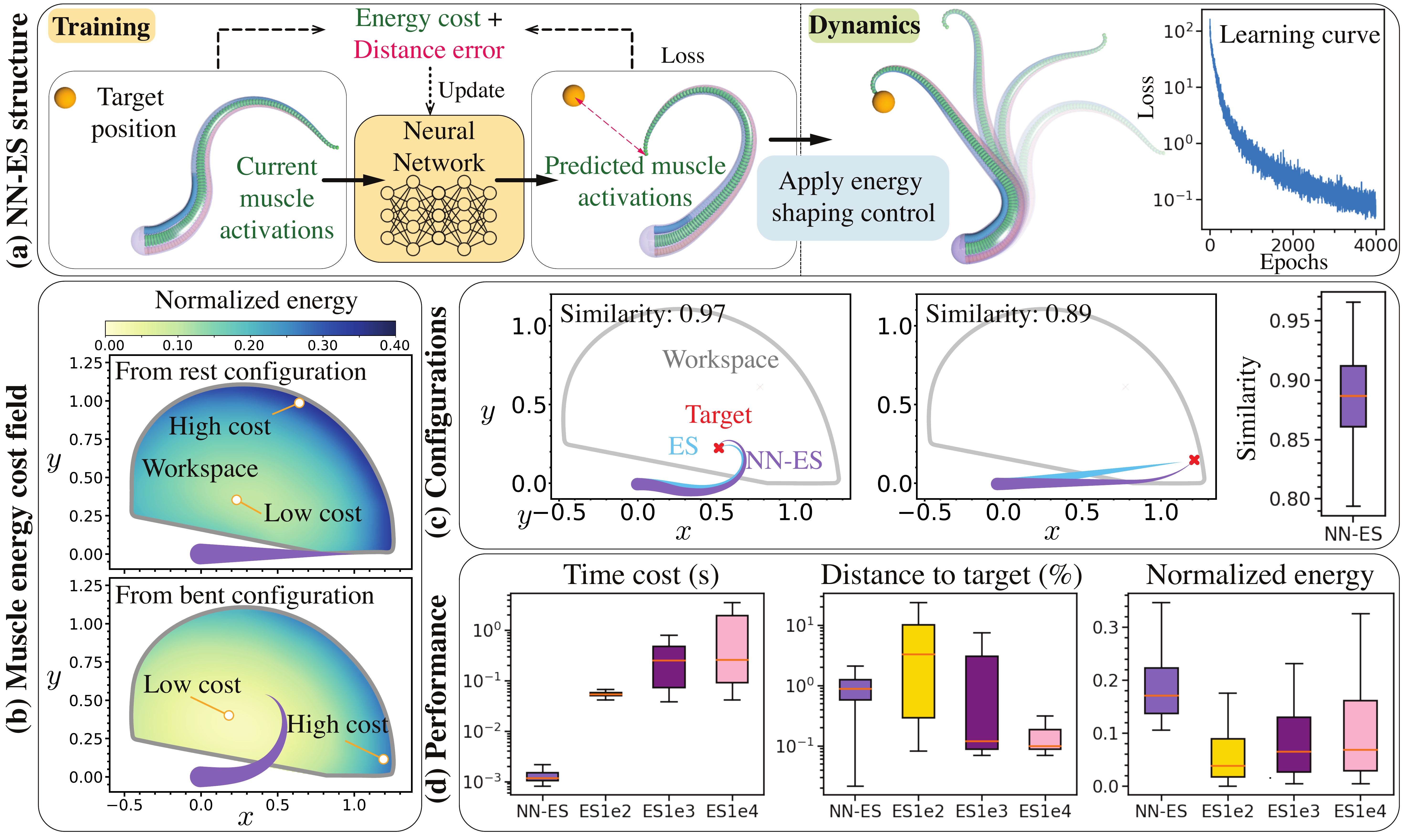}
    \captionsetup{belowskip=-10pt}
    \caption{
    Arm-level controller: (a) Neural Network Energy Shaping (NN-ES) control utilizes a learned mapping to determine static muscle activations, and dynamically brings the arm to a given food target. The mapping, represented as a neural network, is trained to take as inputs the food target location and current muscle activations, and then outputs muscle activations that minimize tip-to-food distance and energy expenditure.
    (b) Muscle energy-cost (Eq.~\ref{eq:ESoptimization}) normalized by arm length shows the cost of NN-ES to reach a point within the workspace 
    $\protect\mathcal{W}$ 
    given the starting arm configuration (top panel--initially straight arm, bottom panel--initially bent arm).
    (c) NN-ES obtained solutions have high similarity relative to iterative ES solutions \cite{Chang:2021}. Differences may arise when targets are located towards the edge of the workspace. Indeed, arm configurations obtained via NN-ES are observed to bend near the tip, while ES solutions bend closer to the base. Right panel shows box plot of similarity score between NN-ES and iterative ES solutions, with orange line representing the median, purple box representing the inter-quartile (middle 50\%) range, and whiskers denoting the min and max of 100 evaluation samples. 
    (d) Performance of NN-ES and iterative ES, the latter considering an increasing number of iterations in termination condition. Orange lines represent the metric's median value, boxes represent the inter-quartile (middle 50\%) range, and whiskers denote the min and max of 100 evaluation samples. 
    Comparison shows NN-ES achieves solutions over 200x faster than the iterative ES scheme, while achieving median tip-to-food distances (normalized by the arm length) of less than 1\%. The median muscle energy cost of iterative ES increases with the max number of iterations allowed. This reflects the fact that obtained solutions improve with more iterations, decreasing the tip's distance to the target, often through bending and thus higher actuation costs. NN-ES has slightly higher median energy cost (and similar maximum energy cost) as the most accuracy iterative ES solution.}
    \label{fig:low-level}
\end{figure*}

\textbf{Fast neural-network energy shaping (NN-ES).} While we previously demonstrated the use of energy shaping (ES) for muscle coordination in a soft arm \cite{Chang:2020_ES,Chang:2021, chang2022energy}, the solution to the above optimization problem is reliant on a computationally expensive forward-backward iterative scheme. Here, in a hierarchical context where energy shaping will be frequently called upon by a high-level controller, fast solutions are instead imperative. In response to this need, we replace the forward-backward scheme with a neural network and directly learn the mapping $\pi : \left\{q, \alpha_0(s)  \right\}  \mapsto \alpha(s) $ that takes initial muscle activations $\alpha_0(s)$ and target location $q$, and outputs the activations $\alpha (s)$ that cause the arm to reach $q$ while minimizing muscle costs (Fig.~\ref{fig:low-level}a).

In the CyberOctopus' arm, muscle activations are continuous over $s\in[0,L_0]$, requiring us to first obtain a finite-dimensional representation of the activations for use with our neural network, which we accomplish via a set of $K$ orthonormal basis functions $\{e_k (s)\}_{k=1}^K$. The procedure for finding this set is described in the SI. In this basis set, the continuous muscle actuation profile $\alpha (s)$ is represented by the coefficients $\{\hat{\alpha}_k\}_{k=1}^K$, so that $\alpha (s) = \sum_k \hat{\alpha}_k e_k (s)$. 
The inputs to the network are then the coefficients of the initial actuation profile $\{\hat{\alpha}_{0,k}\}$ along with the target location $q$. Denoting the network weights as $\mathsf{v}$, the outputs of the network are then the coefficients of the desired muscle activation profile $\{\hat{\alpha}_k (\mathsf{v})\}$. The loss function of the network is then obtained by recouching Eq.~\ref{eq:ESoptimization} as a function of the network weights $\mathsf{v}$
\begin{align}
\label{eq:loss_function}
\begin{array}{c}
\text{Neural-Network} \\ \text{Energy Shaping} \\ \text{(NN-ES)} 
\end{array}
\left\{\rule{0cm}{0.5cm}\right. %\quad
		\minimize_{\mathsf{v}} ~ J \left( \sum_{k=1}^{K} \hat{\alpha}_{k} (\mathsf{v}) e_k(s); \sum_{k=1}^{K} \hat{\alpha}_{0,k} e_k (s), q \right)
\end{align}
Individual muscles' activation bounds ($\alpha^\muscle(s) \in [0,1]$) of Eq.~(\ref{eq:ESoptimization}) are enforced via the function $\max(\min(\alpha^\muscle(s),0),1)$.
% accounted for here by clipping muscle activations at the boundaries (0 or 1) if the network output is outside the allowable range $[0,1]$.

%%%%%%%%%%%%%%%%%%%%%%%%%%%%%%%%%%%%%%%%%%%%%%%%%%%%%%%%%%%%%%%%%%%%%%%
%%%%%%%%%%%%%%%%%%%%%%%%%%%%%%%%%%%%%%%%%%%%%%%%%%%%%%%%%%%%%%%%%%%%%%%

\subsection{An arm reaching for food}
To enable our arm, we train the mapping $\pi$, represented as a feedforward neural network with three hidden layers of 128 Rectified Linear Unit (ReLU) activation functions.
This process can be summarized as follows.
The network is trained for 4000 epochs. 
For each epoch, 100 training samples are generated with initial activations $\alpha_0(s)$ randomly selected from a Gaussian distribution, and target locations $q$ (food) randomly selected from a uniform distribution over the workspace $\mathcal{W}$ (the set of all points reachable by the tip of the arm).
For each training sample, the neural network produces an $\alpha(s,\mathsf{v})$, from which the target-to-tip distance $\delta(\alpha(s,\mathsf{v}),q)$ is computed based on the resulting equilibrium configuration (Eq.~\ref{eq:matchingcondition} and Eq.~\ref{eq:state_kinematics}). Because $\delta$ directly depends on the neural network weights $\mathsf{v}$ --- through $\alpha(s,\mathsf{v})$ --- we can compute the gradient of Eq.~\ref{eq:loss_function} with respect to $\mathsf{v}$ and thus update $\pi$ in an unsupervised manner.

 As seen in Fig.~\ref{fig:low-level}a, the network successfully learns, minimizing the loss function of Eq.~\ref{eq:loss_function}. 
Exploring the characteristics of the learned mapping, we find that the initial configuration of the arm plays a substantial role in determining muscle activation costs (Fig.~\ref{fig:low-level}b). 
For a straight initial configuration ($| \alpha_0(s) |^2 =0$), targets in the middle of the workspace require less change in muscle activation. 
Indeed, the arm can reach these targets by activating only the longitudinal muscles along the arm's distal end, which is thinner and hence less stiff. 
In contrast, targets at the boundary of $\mathcal{W}$ require recruiting both longitudinal and transverse muscles to bend the base (thick and stiff) and extend the arm. 
For a bent initial configuration, the change in muscle energy %$\lVert \alpha(s, \mathsf{v}) - \alpha_0(s)\rVert^2$ 
is generally lower since longitudinal muscles are already partially activated. This is particularly true for reaching the center of the workspace. 

We next proceed to compare our NN-ES approach with the original iterative ES \cite{Chang:2021}, employing the termination conditions of normalized tip-to-target distance $\delta(\alpha,q)/L_0 < 0.01\%$ or 10,000 maximum iterations.
To quantify differences in the obtained equilibrium configurations $x(s)$, we introduce the similarity metric $D=\left(1+\int_0^{L_0}\left| x^\text{NN-ES}(s)-x^\text{ES}(s)\right| \dif s\right)^{-1} \in (0,1]$, where 1 indicates identical solutions.
As seen in Fig.~\ref{fig:low-level}c, for 100 randomly generated cases, NN-ES and ES produce solutions characterized by a high degree of similarity (average $D=0.89$). 
Differences appear for targets located far from the base, with NN-ES stretching and bending the arm towards the tip, while ES directly orients the entire arm in the food direction. 
Both algorithms are accurate in reaching food, achieving median tip-to-target distances of less than 1\% relative to the rest length of the arm, although we note that NN-ES tends to utilize slightly larger muscle activations than ES (Fig.~\ref{fig:low-level}d).
This drawback is nonetheless compensated by a significant reduction in solution time (Fig.~\ref{fig:low-level}d), whereby NN-ES outperforms ES by a factor of 200. Further, we note that while ES performance may depend on the allowed maximum number of iterations, the trends described above persist as we span from 100 to 10,000 max iterations. 
Taken together, these results demonstrate NN-ES to be fast and accurate in coordinating muscle activity and in executing low-level reaching motions. We thus conclude that NN-ES is suitable to be integrated into our hierarchical approach.

\section{Central-level problem: coordinating foraging behavior}\label{sec:cyberoctopus_overview}

We now turn to the problem of a CyberOctopus foraging for food within a two-dimensional (planar) arena (Fig.~\ref{fig:problem}b). Inspired by real octopuses coordinating their arms to move and collect food \cite{Kennedy:2020, levy2017motor}, the CyberOctopus is tasked with maximizing the energy intake derived from collecting food, while minimizing the muscle activations required to reach for it. By engaging its multiple arms, the CyberOctopus can move in any planar direction without re-orienting its body. This multi-directionality, compounded by the difficulties associated with distributed muscular actuations across multiple arms, muscle expenditure estimation, limited workspace, and potential presence of solid obstacles, renders the foraging problem challenging.

Here, we define the behavioral primitives available to the central-level controller for orchestrating foraging behavior (Sec.~\ref{sec:behavioral_primitives}), before providing the full problem's mathematical formulation (Sec.~\ref{sec:foraging_general}). 
Finally, we describe a reinforcement learning-based approach (Sec.~\ref{sec:learning-solution}) that utilizes a spatial attention strategy to simplify the planning process, allowing us to successfully control the CyberOctopus. 

\subsection{Behavioral primitives: \textit{reach all} and \textit{crawl}}\label{sec:behavioral_primitives}
The combination of low-level motor primitives into ordered sequences allows us to construct basic behavioral primitives. We define two behavioral primitives, \textit{reach all} and \textit{crawl}, in an attempt of abstracting out the complexity of foraging into simple terms.
Nonetheless, our decomposition approach conveniently provides the opportunity and freedom of defining arbitrary command sets, and different choices may be made.

% Behavioral primitives consist of sequences of motor primitives applied to an individual arm.  %$\alpha[\hat{n}]$ 
 %, where $\hat{n}$ denotes the sequence order. For example, a $\textit{reach all}$ primitive would result in the activation sequence $\{\alpha[\hat{n}=1],\alpha[\hat{n}=2],...\}$ for reaching all available food targets, as described below. 

The \textit{reach all} behavior consists in the arm attempting to reach all food targets within its workspace $\mathcal{W}$.
Food is collected sequentially with the ordering determined in a greedy manner. 
The arm calculates --- via the NN-ES controller --- the change in muscle activation needed to collect each food target from its current configuration, and then collects the target that requires the least change. This process is repeated until all food in $\mathcal{W}$ is collected (or attempted to).

The \textit{crawl} behavior consists of a predefined set of muscle activations $\alpha^{\text{TM}}_{\text{crawl}}$. 
First, transverse muscles are activated to extend the arm horizontally by a fixed amount $\Delta r$ along the crawling substrate. After extension, suckers at the tip of the arm engage with the substrate and transverse muscles are relaxed, pulling the octopus forward by the amount $\Delta r$, at which point the suckers are released. We note that even though we do not explicitly model the suckers, their effect is accounted for by appropriately choosing the arm boundary conditions.

% \begin{figure}[t!]
% \centering
% \includegraphics[width=0.35\linewidth]{figures/grid.pdf} 
% \caption{Setup of the foraging problem showing the CyberOctopus surrounded by food targets arranged on a grid that matches the distance traveled during one \textit{crawl} step. The yellow dome shows the attention space heuristic described in Sec.~\ref{sec:learning-solution} where the CyberOctopus only considers food items within this space. }
% \label{fig:wrapfig}
% \end{figure}
%%%%%%%%%%%%%%%%%%%%%%%%%%%%%%%%%%%%%%%%%%%%%%%%%%%%%%%%%%%%%%%%%%%%%%%%%%%%%%%%%%%%%%%%%%%%%%%%
\subsection{Mathematical formulation of the foraging problem in hierarchical form} \label{sec:foraging_general}
\begin{figure*}[t!]
\centering
    \centering
\includegraphics[width=\textwidth]{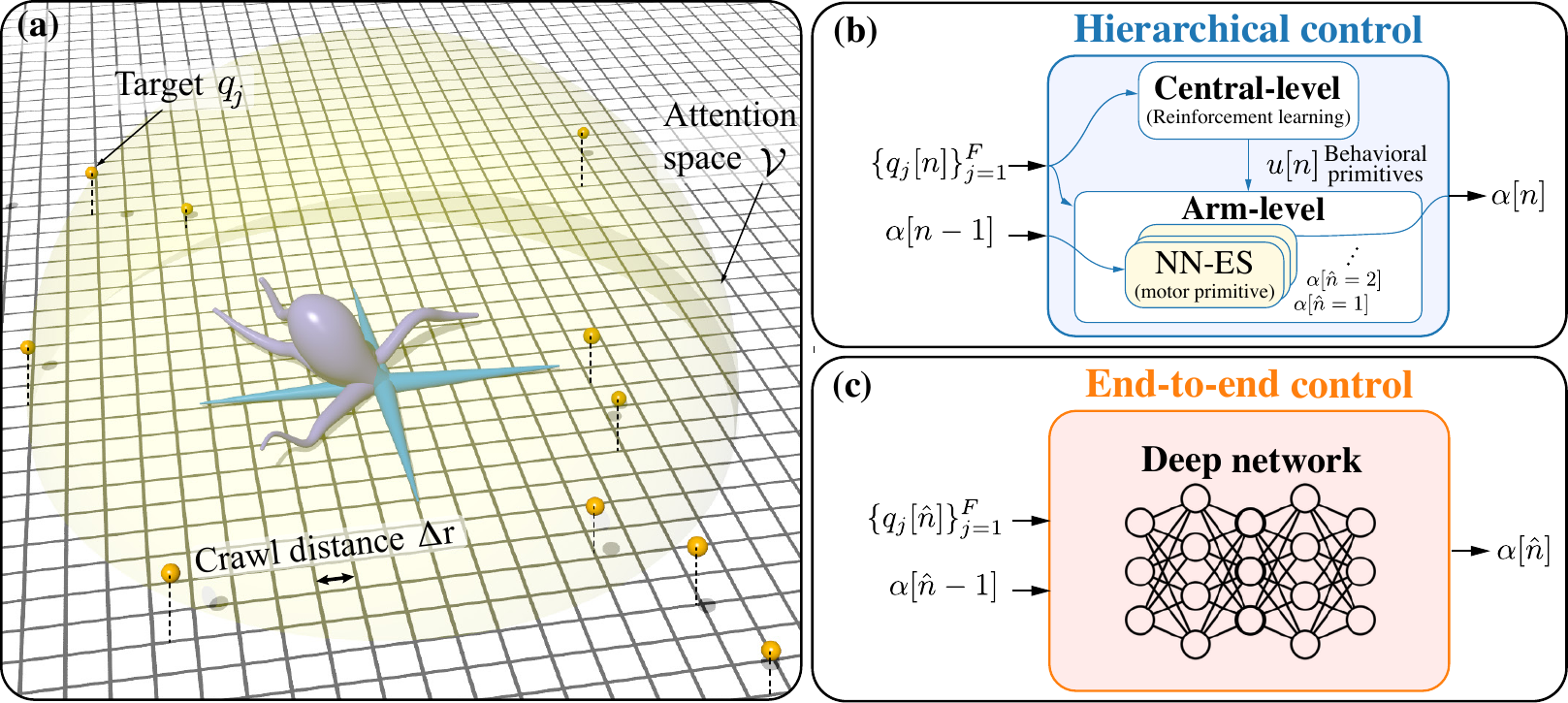}
\captionsetup{belowskip=-10pt}
\caption{(a) Setup of the foraging problem showing the CyberOctopus surrounded by food targets arranged on a grid whose spacing matches the distance traveled during one \textit{crawl} step. The yellow dome shows the attention space heuristic described in Sec.~\ref{sec:learning-solution}, so that the CyberOctopus only considers food items within this space. 
(b) Block diagram for hierarchical controller (Sec.~\ref{sec:foraging_general},~\ref{sec:learning-solution}) and 
(c) end-to-end controller (whose mathematical formulation can be found in the SI).
Both approaches make use of the spatial attention heuristic, and receive the same inputs to produce one activation output.
However, the hierarchical decomposition allows us to rewire internally the flows of information, separating concerns into qualitatively different tasks, which can then be efficiently solved by appropriate algorithms.
}
\label{fig:grid_block}
\end{figure*}

Let us consider a CyberOctopus with $I$ active arms aiming to collect $T$ food items that are scattered randomly throughout an arena. Food can be found at any vertical location, while the horizontal coordinates are constrained to a two-dimensional Cartesian grid formed by discrete crawling steps (Fig.~\ref{fig:grid_block}a). With respect to the bases of the arms, the location of the $j$-th target is denoted as $q_j \in \R^3$. If the CyberOctopus has complete knowledge of all $T$ food item's locations, the problem for the CyberOctopus is to create an optimal plan that sequentially composes the behavioral primitives --- \textit{reach all} and \textit{crawl} --- for all of the $I$ arms, so as to fetch all of the food, while minimizing muscle energy expenditure. 

Before we model the full problem mathematically, it is useful to examine the simplest case of only one active arm where all targets are within the arm's workspace $\mathcal{W}$. In this case, the CyberOctopus can simply execute a single \textit{reach all} behavior to gather all the targets. However, in cases where some of the targets lay outside $\mathcal{W}$, the CyberOctopus will also need to \textit{crawl}. Depending on the location of the targets, it may be beneficial for the CyberOctopus to \textit{crawl} even if there are already targets in its workspace. Doing so may serve to bring additional targets within reach, to gather them all more efficiently with a single \textit{reach all} maneuver.

We formulate this optimal planning problem as a Markov Decision Process (MDP).    
Even though time has been continuous so far, for the MDP model we consider a discrete-time formulation. Temporal discretization naturally arises by considering the time elapsed between the start and the end of a primitive. Since the motor primitives are nested inside a behavioral primitive (as described in Section \ref{sec:behavioral_primitives} and Fig.~\ref{fig:grid_block}b), we introduce two different discrete-time indices, at the behavioral ($n$) and at the motor primitive level ($\hat{n}$).

% Thus, at the behavioral level, the discrete time is denoted as $n=0,1,2,...$. 
Thus, at time $n$, each arm executes a behavioral primitive (action) $u_i[n],~i=1,..., I$, which takes value in the set $\{\textit{reach all}, \textit{crawl}\}$. 
This means that each individual arm is treated as functionally equivalent, reflecting the observed bilateral symmetry of octopus arms \cite{Bidel:2022}, and tries to either \textit{reach all} available food targets in its workspace $\mathcal{W}_i$, or \textit{crawls} a fixed amount $d_i$ along its own direction (where $|d_i| = \Delta r$). As a consequence, multiple reaching planes become simultaneously accessible, rendering the foraging problem three-dimensional. Taking the actions of all arms together, the decision variable at time $n$ is denoted as $u[n] = \{u_i[n]\}_{i=1}^I$. 

The state at time $n$ is denoted as $z[n]$.  Its components are
\begin{align}
z[n]= \{q_j[n],f_j[n]\}_{j=1}^T 
\label{eq:state_single_arm}
\end{align}
where the flag $f_{j}[n]\in\{0,1\}$ signifies whether the $j$-th target
has already been collected ($f_j [n] = 0$)  or not ($f_j [n] = 1$).  
%The state-space is denoted as $\mathcal{Z}$.  
The dynamics then become
\begin{align}
    \begin{split}
        q_j [n+1] &= q_j [n]-\sum_{i=1}^I d_i\mathds{1}(u_i[n]=\textit{crawl})\\
        f_j[n+1] 
        &=  f_j [n] \prod_{i=1}^I (1-\mathds{1}(u_i[n]=\textit{reach all} \ \& \ q_j\in\mathcal{W}_i))\\
        % f_j [n] \prod_{i:q_j\in\mathcal{W}_i,i\in\mathcal{I}}\mathds{1}(u_i[n]=\textit{crawl})
    \end{split}
  \label{eq:state_dynamics_multiple}
\end{align}
where $\mathds{1}(\cdot)$ is the indicator function. Here the first equation expresses the change in the positions (relative to the arm base) of food targets after a \textit{crawl} step, while the second term denotes the change in collection status of food targets after a \textit{reach all} step. 
When executing the selected behavioral primitives, the CyberOctopus first reaches with any arm that selected \textit{reach all}, before performing any crawling. Finally, the CyberOctopus is not allowed to crawl through the boundaries of the arena. If a boundary is approached, the CyberOctopus remains in place unless it chooses to crawl alongside or backtracks.

%%%%%%%%%%%%%%%%%%%%%%%%%%%%%%%%%%%%%%%%%%%%%%%%%%%%%%%%%%%%%%%%%%%%%%%%%%%%%%%%%%%%%%%%%%%%%%%%
%%%%%%%%%%%%%%%%%%%%%%%%%%%%%%%%%%%%%%%%%%%%%%%%%%%%%%%%%%%%%%%%%%%%%%%%%%%%%%%%%%%%%%%%%%%%%%%%

% \medskip
Subject to the dynamics of Eq.~\ref{eq:state_dynamics_multiple}, the CyberOctopus aims to find the optimal sequence of behavioral primitives $\bar{u}:=\{u[0],u[1],u[2],\hdots,u[N-1]\}$ so as to maximize the cumulative reward
\begin{align}
\begin{split}
\max\limits_{\bar{u}}~~  J(\bar{u}) &= \sum_{n=0}^{N-1} R(z[n],u[n]) \\
\text{subject to~~~} & \quad\text{the dynamics (Eq.~\ref{eq:state_dynamics_multiple}) and given}~ z[0] 
\end{split}
\label{eq:MDP}
\end{align}
where $N$ is a given stopping time.

The reward function is chosen to capture the trade-off between negative energy expenditure associated with muscle activation, and positive energy associated with food collection. It is defined as
\begin{align}
\label{eq:multi_arm_reward}
R(z[n],u[n]) = -\Phi[n] + \sum_{i=1}^I \begin{cases}
- E^{\text{c}} \hspace{78pt}\text{if $u_i[n]=$ \textit{crawl}}\\
- E^{\text{r}}_i [n] + \gamma f_{\mathcal{W}_i}[n] \hspace{16pt}\text{if $u_i[n] =$ \textit{reach all}}
\end{cases}
\end{align}
where $E^{\text{c}}$ and $E^{\text{r}}_i[n]$ are the total muscle activation costs of the multiple low-level motor primitives necessary to complete the selected command $u_i[n]$ for the $i$-th arm, $\gamma$ is the energy of an individual food target,
$f_{\mathcal{W}_i}[n]$ is the total number of food items collected during a \textit{reach all} execution by the $i$-th arm, and $\Phi[{n}]$ is a penalty term if all $I$ active arms choose \textit{reach all} when no food is available. 

If $u_i [n] = \textit{reach all}$ is selected, then the muscle activation sequence $\{\alpha_i[\hat{n}=1],\alpha_i[\hat{n}=2],...,\alpha_i[\hat{n}= \hat{f}_{\mathcal{W}_i}[n]]\}$ is generated for reaching all food targets in the workspace $\mathcal{W}_i$ (as described in Section \ref{sec:behavioral_primitives}, where $\hat{n}$ denotes the discrete-time at the motor level, and $\hat{f}_{\mathcal{W}_i}[n]$ is the total number of food items within reach at time $n$).
If instead $u_i[n] = \textit{crawl}$, then the predefined activation $\alpha^{\text{TM}}_\text{crawl}$ is recruited.
Corresponding muscle activation costs are defined as
\begin{align}
\begin{split}
  E^{\text{c}} &= \frac{1}{L_0}\int_0^{L_0} \abs{\alpha^{\text{TM}}_\text{crawl}}^2 \dif s \\
E^{\text{r}}_i[n] &=\frac{1}{L_0}\int_0^{L_0} \sum_{\hat{n}=1}^{\hat{f}_{\mathcal{W}_i}[n]}\abs{\Delta\alpha_i[\hat{n}]}^2 \dif s
\end{split}
 \label{eq:Ereach}
\end{align}
where 
$\Delta\alpha_i[\hat{n}]=\max\{\alpha_i[\hat{n}] - \alpha_i[\hat{n}-1],0\}$
is the difference between successive muscle activations. The maximum operator is used so that only actions leading to increased muscle activation are accounted for, i.e. there is no cost for relaxing muscles. If the low-level controller collects all food targets within reach (which is generally true if no obstacles are present) then  ${f}_{\mathcal{W}_i}[n] = \hat{f}_{\mathcal{W}_i}[n]$.

\subsection{Spatial attention heuristic and Reinforcement Learning solution method} \label{sec:learning-solution}
In general, the high-level problem of Eq.~\ref{eq:MDP} has no analytical solution. We then resort to searching for an approximate one, incorporating further insights from the octopus' behavior. 
Octopuses integrate visual, tactile, and chemo-sensory information to forage and hunt. However, in the wild they are thought to primarily rely on visual cues \cite{Maselli:2020}. For example, during foraging, they make behavioral decisions based on their distance from the den \cite{Mather:1991}, and they are able to discriminate between objects based on size, shape, brightness, and location \cite{Boal:1996}. These observations suggest the potential use of target prioritization strategies based on spatial location. 

Adopting this insight, we define a cognitive attention heuristic wherein the CyberOctopus only pays attention to uncollected food targets that are within the attention space $\mathcal{V}$, ignoring all others. This attention space $\mathcal{V}$ can be flexibly defined depending on the task at hand (see SI for details). Here, we use an attention space that extends out twice the workspace distance (Fig.~\ref{fig:grid_block}a).
The CyberOctopus' cognitive load can be further relieved by considering a fixed, maximum number of closest targets, allowing for immediate planning, while retaining sufficient environmental awareness for adequate longer-term decision-making.
With this heuristic, the state of Eq.~\ref{eq:state_single_arm} reduces to
\begin{equation}
    z[n] = \{q_j[n]\}_{j=1}^{F}
    \label{eq:RL_state}
\end{equation}
where $\{q_j[{n}]\}_{j=1}^{F}$ are the positions (relative to the arm base) of the $F$ closest uncollected food targets within $\mathcal{V}$. If fewer than $F$ targets are within the attention range, the excluded entries are set to 0. If more than $F$ targets are within the attention range, only the first $F$ are considered. This state definition is the one employed throughout the remainder of the paper and is the one depicted in Fig.~\ref{fig:grid_block}b.

The use of this heuristic leads to a fixed state-space size, making the problem naturally amenable to reinforcement learning approaches. Here we employ the Proximal Policy Optimization (PPO) algorithm \cite{Schulman2017}, considered to be a state-of-the-art on-policy reinforcement learning scheme due to its robust performance. PPO utilizes an actor-critic architecture where an actor network encodes the control policy, while a critic network estimates the quality of the control policy. Throughout the rest of the paper, the control policy is encoded in a feedforward neural network with three hidden layers ($32\times32\times16$) of ReLU activation functions. The critic network shares the first two hidden layers with the control policy network but has a separate third hidden layer, also with 16 neurons.

%%%%%%%%%%%%%%%%%%%%%%%%%%%%%%%%%%%%%%%%%%%%%%%%%%%%%%%%%%%%%%%%%%%%%%%%%%%%%%%%%%%%%%%%%%%%%%%%
\begin{figure*}[t!]
\centering
    \centering
\includegraphics[width=\textwidth]{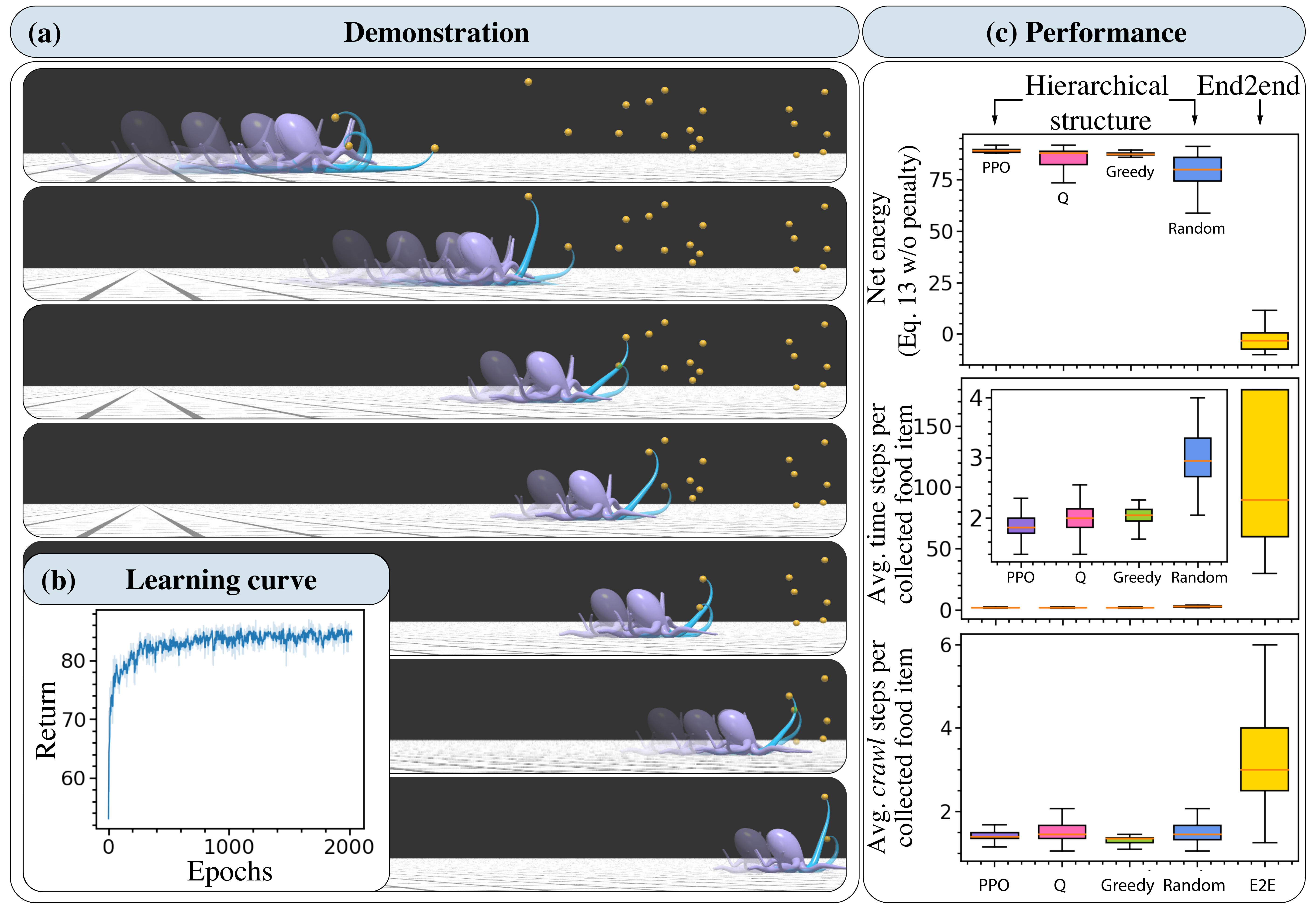}
\captionsetup{belowskip=-10pt}
\caption{Hierarchical control framework. Based on the locations (relative to the arm base) of targets in the view range, the high-level controller decides between two actions, \textit{crawl} or \textit{reach all}, to maximize gained energy. 
(a) Demonstration of learned PPO policy for a single active arm as implemented in \textit{Elastica}. The active arm is depicted in blue, with the levels of shading indicating intermediate configurations. The CyberOctopus successfully moves forward and collects available food. Here, $F=5$, $\gamma=5$, and $\Phi=1.0$. A video of this demonstration is available (SI Video 1).
(b) Learning curve of the PPO algorithm for a single active arm. 
(c) Key metrics for evaluating the performance of hierarchical and end-to-end (e2e) control schemes. 
Orange lines represent the metric's median value, boxes represent the inter-quartile (middle 50\%) range, and whiskers denote the min and max of 100 evaluation samples. 
PPO is the reinforcement learning algorithm, Q is the analytic solution of a simplified DP problem (see SI for details), the greedy policy chooses to reach anytime there is food within reach, else it crawls, and the random policy chooses actions randomly. 
The e2e approach attempts to directly solve the end-to-end problem (see SI for details).
Hierarchy-based control policies are found to outperform end-to-end solutions, with PPO performing best overall. 
% \textcolor{red}{ToDo: Define F and other key values.}
}
\label{fig:demo}
\end{figure*}
%%%%%%%%%%%%%%%%%%%%%%%%%%%%%%%%%%%%%%%%%%%%%%%%%%%%%%%%%%%%%%%%%%%%%%%%%%%%%%%%%%%%%%%%%%%%%%%%

% \textcolor{red}{[Should we make a 'Section 5: Results' here?]}
\section{A CyberOctopus foraging for food}
Next, we put to the test the combined machinery described in Sections \ref{sec:Primitives} and \ref{sec:cyberoctopus_overview} by simulating and characterizing a CyberOctopus foraging for food.
To illustrate the use of primitives for high-level planning and reasoning, we first consider the reduced problem of a single active arm (Sec.~\ref{sec:foraging_single}), for which an analytical solution can be obtained under simplifying assumptions. After showing favorable comparison between learning-based and analytical solutions, in Sec.~\ref{sec:foraging_multiple} we expand our approach to the case of multiple active arms (up to four), for which no analytical solutions exist. Finally, a third (lowest) level of control based on local physical compliance is incorporated, and a CyberOctopus outfitted with this tri-level hierarchy is shown to forage in an arena littered with obstacles in 3D space.

\subsection{Foraging with one arm} \label{sec:foraging_single}
\hspace{\parindent}\textbf{Analytical solution.} As mentioned aboved, there is no analytical solution to the high-level problem of Eq.~\ref{eq:MDP}. However, under simplifying assumptions, analytical solutions may be obtained for the case of a single arm ($I = 1$). Here, we make the following assumptions: (1) The workspace $\mathcal{W}$ is treated as a rectangle whose width is larger than the distance the arm can crawl in one step; 
(2) The energetic cost of each high-level command is simplified to be a constant, with no dependence on the NN-ES muscle activations;
% i.e. ${E}_{\text{c}}$ and ${E}_{\text{r}}$ are constants. 
(3) The food reward $\gamma$ is greater than the constant cost to reach a target. %such that $\gamma >\mathcal{E}_\text{reach}$. 
Under these assumptions, an analytical Q-policy solution to the optimal dynamic programming (DP) planning problem of a CyberOctopus foraging with one arm can be derived, as detailed in the SI. 
We note that the same procedure can be extended to two arms crawling and reaching in two orthogonal planes (SI). However, if more than two arms are considered, the number of steps to reach all food targets can not be determined, making the derivation of an optimal analytical policy not possible, even under the above simplifying assumptions. However, despite their limited scope, one- and two-arm analytical solutions are still useful to benchmark and contextualize our hierarchical approach and its learning-based solution, as we progress to the more general, non-analytically tractable scenario of larger numbers of engaged arms. 

\textbf{Reinforcement Learning solution.} We proceed with solving the problem of Eqs.~\ref{eq:state_single_arm}-\ref{eq:RL_state} for one arm, via the PPO reinforcement learning algorithm, using the setup described in Section \ref{sec:learning-solution}. The performance of the one-arm policy, numerically obtained with PPO and dynamically executed by the CyberOctopus in \textit{Elastica}, is shown in Figure \ref{fig:demo}a, while policy convergence during training is illustrated in Fig.~\ref{fig:demo}b. 
The policy is trained for 2000 epochs, each epoch entailing 1024 time steps. At the beginning of each episode, 20 targets are randomly generated in a rectangular area within the arm's bending plane (Fig.~\ref{fig:problem}b), and the arm is initialized in a straight configuration on the left side of the food (Fig.~\ref{fig:demo}a). An episode terminates when either all targets have been reached or the number of time steps exceeds 180, so that each epoch contains at least 5 episodes.

The CyberOctopus (whose only active arm is depicted in blue) successfully learns to crawl towards food, positioning itself for reaching until all food in the environment is collected (Fig.~\ref{fig:demo}a and SI Video 1).
To contextualize this performance, we implement three alternative high-level controllers: the simplified, analytical Q-policy solution described above, a `greedy' controller that immediately reaches for food whenever available, otherwise it chooses to crawl, and a random controller that selects \textit{crawl} or \textit{reach all} with equal probability at each time step.
We evaluate the performance of these four controllers on three metrics (Fig.~\ref{fig:demo}c): (1) net energy (Eq.~\ref{eq:MDP}), (2) average number of time steps per food item collected, and (3) average number of crawl steps per food target.
We find, unsurprisingly, that the random policy performs the worst, utilizing, on average, 25\% more energy and 40\% more steps per episode than the other approaches. 
The greedy and Q-policy controllers perform comparably, though the Q-policy presents a wider distribution than greedy, likely due to simplifying assumptions that cause the occasional miss of a target. 
Overall, the PPO policy exhibits the best performance, strategically crawling until a large number of food items are simultaneously within reach, to then fetch them all at once. In light of the energy cost maps of Fig.~\ref{fig:low-level}d, this learned approach intuitively correlates to lower muscle activation costs. 
This strategy (which is also adopted by the Q-policy in the simplified setting of the problem) not only reduces energy costs but additionally allows completion of the foraging task in a fewer number of time steps (Fig.~\ref{fig:demo}c). As we will see, the performance gap between PPO and the alternative approaches will only widen as more complex scenarios ($I>1$) are considered, until becoming the only reliable option. 

To comparatively isolate the benefits provided by our hierarchical decomposition, we solve the same foraging problem in an end-to-end (e2e) fashion, using PPO. In other words, we train a single network to directly map food locations and current muscle activations to output muscle activations, completely bypassing the decomposition between high-level planning and low-level execution.
A comparison between the two approaches is schematically illustrated through the block diagrams of Fig.~\ref{fig:grid_block}b,c, with further mathematical and implementation details available in the SI.
We note that while the previous activation $\alpha[\hat{n}-1]$ is used by the low-level controller in Fig.~\ref{fig:grid_block}b, in the end-to-end formulation depicted in Fig.~\ref{fig:grid_block}c, this information is included in the state representation. 
Thus, overall, the same amount of information is provided to both frameworks. 

We find that all four hierarchical policies (including the random policy) outperform the end-to-end approach, across all metrics (Fig.~\ref{fig:demo}c), despite significant effort in tuning training parameters.
This result underscores how the separation of concerns enabled by the hierarchy significantly simplifies control, and thus learning. Further, it illustrates the potential of a mixed-modes approach, where model-free and model-based algorithms are employed at different levels so as to synergize and complement each other.

\subsection{Foraging with multiple arms} \label{sec:foraging_multiple}
We next gradually increase the number of active arms engaged by the CyberOctopus, considering first two arms and then four arms. As the number of arms increases, the CyberOctopus is able to access additional directions of movement and must coordinate its arm's behaviors, significantly increasing the problem difficulty. Additionally, in the four-arm case, we incorporate obstacles into the arena, requiring the arms to exploit their mechanical intelligence to avoid becoming obstructed. 
%%%%%%%%%%%%%%%%%%%%%%%%%%%%%%%%%%%%%%%%%%%%%%%%%%%%%%%%%%%%%%%%%%%%%%%%%%%%%%%%%%%%%%%%%%%%%%%%
\begin{figure*}[t!]
\centering
\includegraphics[width=\textwidth]{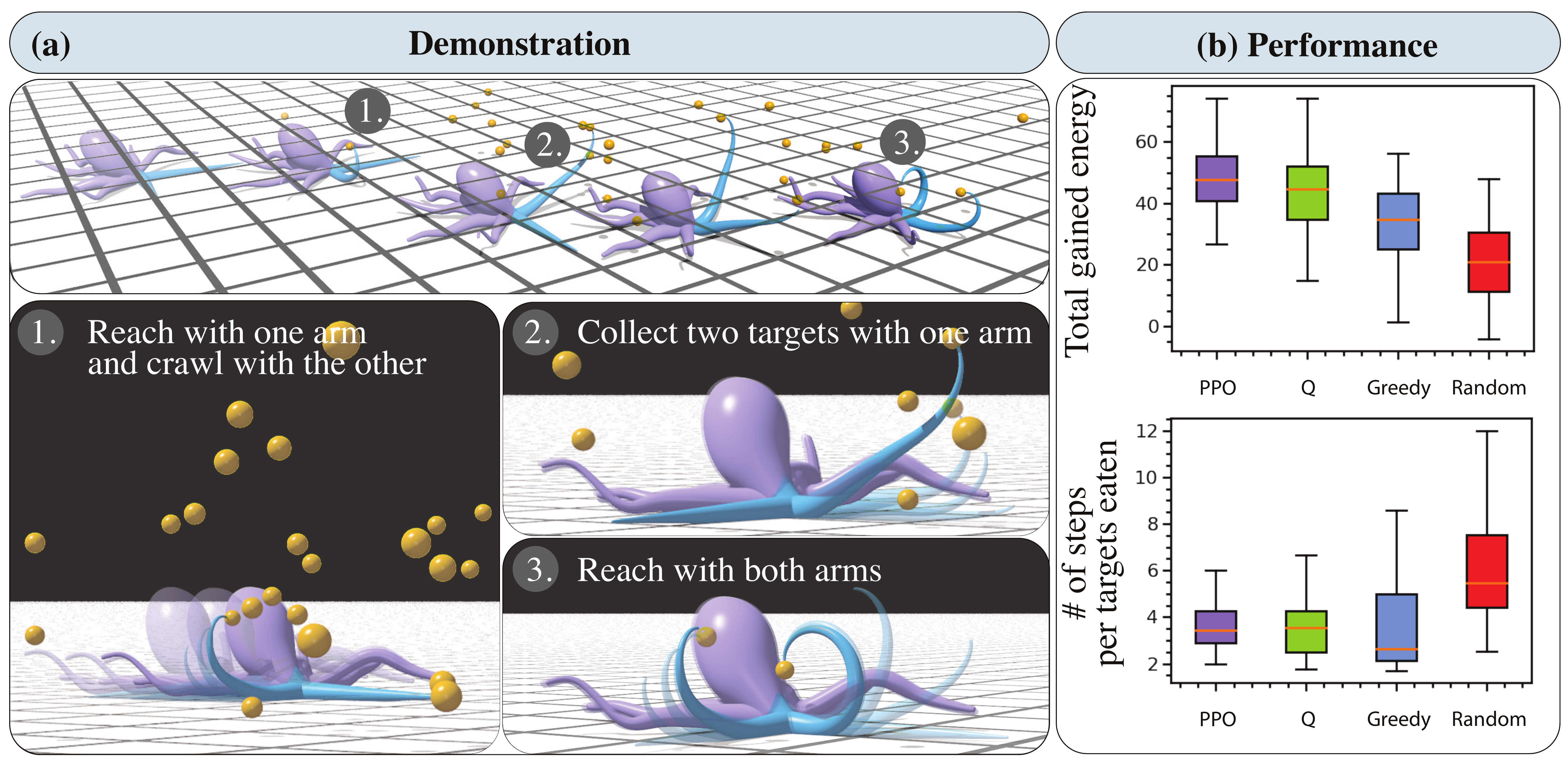}
\captionsetup{belowskip=-10pt}
\caption{(a) Demonstration in \textit{Elastica} of learning-based PPO policy controlling two arms to forage. (a1) The arms coordinate their actions to move across two dimensions and reach targets. (a2) One arm collects multiple targets within its bending plane. (a3) Two arms simultaneously collect targets in their respective bending planes. A video of this demonstration is available (SI Video 2).
(b) Performance of the different control schemes. 
Orange lines represent the metric's median value, boxes represent the inter-quartile (middle 50\%) range, and whiskers denote the min and max of 100 evaluation samples. 
The learning-based PPO policy gains more total energy and requires fewer steps per collected target than the alternative high-level policies. For the learning-based policy, $F=10$. 
}
\label{fig:2arm-demo}
\vspace{-5pt}
\end{figure*}
%%%%%%%%%%%%%%%%%%%%%%%%%%%%%%%%%%%%%%%%%%%%%%%%%%%%%%%%%%%%%%%%%%%%%%%%%%%%%%%%%%%%%%%%%%%%%%%%

\hspace{\parindent}\textbf{Foraging with two arms.} We first consider a CyberOctopus with two active arms ($I=2$), orthogonal to each other. 
Training proceeds as with the single arm, although now 10,000 epochs are used. For each episode, arms are initialized at rest and 20 food locations are scattered randomly throughout the arena on vertical planes that align with the grid formed by discrete crawling steps. As reported in Fig.~\ref{fig:2arm-demo}, the learning-based approach successfully learns and converges (see SI for learning curves). 
% While the centralized approach exhibits better initial learning (see SI for learning curves), it quickly converges to a performance level that the decentralized approach eventually surpasses. 
This is reflected in the evaluation of the corresponding policies in Fig.~\ref{fig:2arm-demo}b.
The behavior learned via the learning-based approach is illustrated in Fig.~\ref{fig:2arm-demo}a as well as in SI Video 2.
As can be seen, the CyberOctopus crawls between targets and collects them, with the learned policy successfully coordinating its two independent arms so as to crawl and switch along orthogonal directions, simultaneously grabbing food with both arms or fetching food with one arm while crawling with the other arm. 
We again contextualize our results by means of three alternative high-level controllers: Q is the analytically identified solution of the simplified DP problem described above and in the SI, 
greedy has each arm collecting food when immediately available, else crawling (SI for details), and random selects \textit{crawl} or \textit{reach all} with equal probability for each arm. 
Again the learning-based solution outperforms the alternative high-level controllers, consistently collecting more energy and using fewer steps to do so (Fig.~\ref{fig:2arm-demo}b). 
We forego a comparison with the end-to-end approach, as its performance is deemed too poor to provide any useful insight. This once again underscores the significant impact of hierarchically decomposing the problem, which in this example amounts to being able to solve the problem versus not being able to do so (end-to-end).

\begin{figure*}[t!]
\centering
\includegraphics[width=\textwidth]{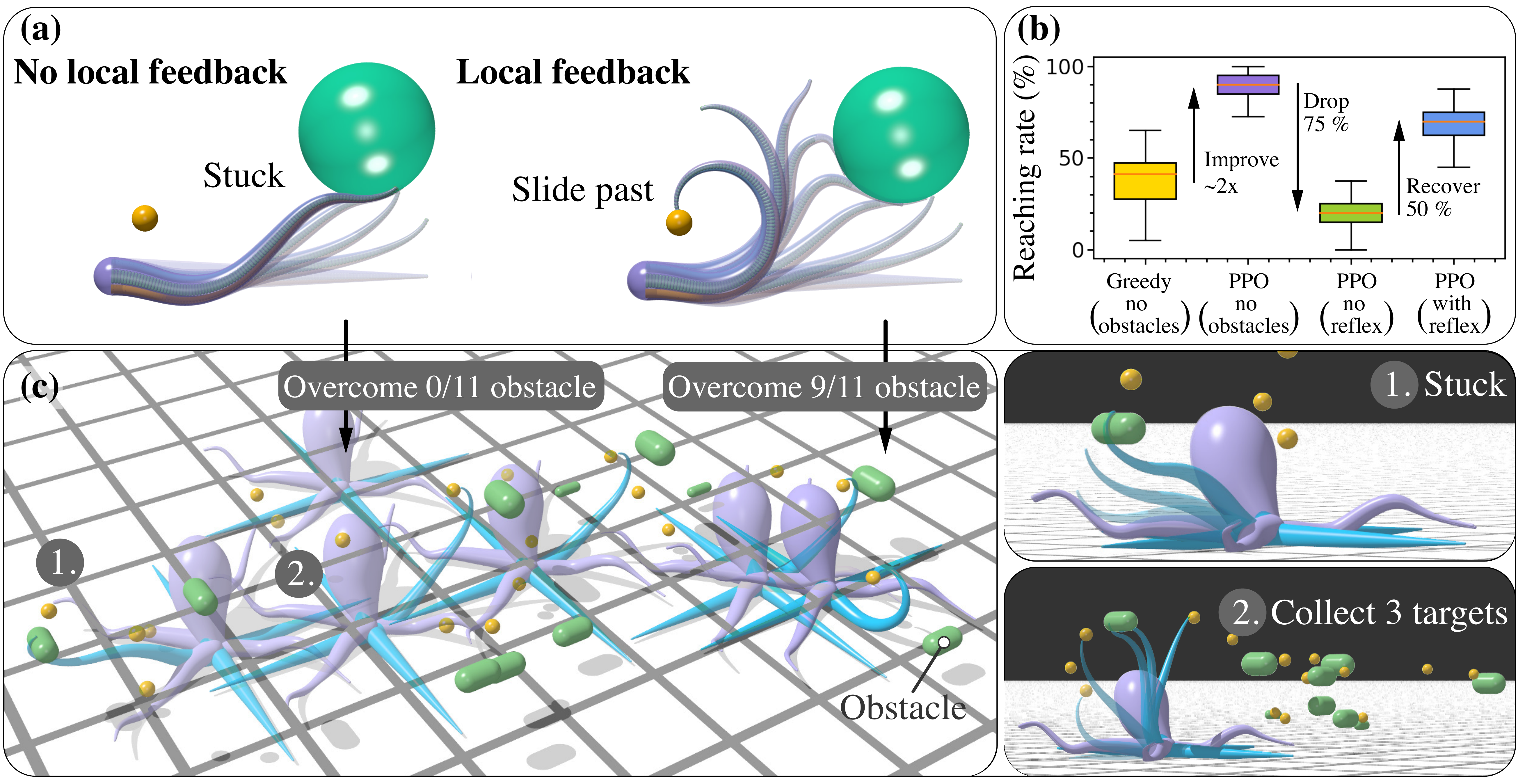}
\captionsetup{belowskip=-10pt}
\caption{
(a) Demonstration of an arm sliding past an obstacle when equipped with local sensory feedback.
(b) Percentage of available food items collected from the environment by the greedy policy (yellow) and the learning-based policy when no obstacles are present (purple), when the arm does not utilize a reflexive behavior (green), and when the reflex is engaged (blue). 
Orange lines represent the median performance over 100 episodes, boxes represent the inter-quartile (middle 50\%) range, and whiskers denote the min and max. 
(c) Demonstration of a foraging CyberOctopus with four active arms (blue). 
Using the local sensory reflex allows the arms to overcome 9 of the 11 obstacles that otherwise would cause the arm to become stuck. 
(c1) The arms coordinate their actions to move across two dimensions and reach targets. (c2) One arm collects multiple targets within its bending plane. 
A video of this demonstration is available (SI Video 3).
}
\label{fig:4arm-demo}
\vspace{-5pt}
\end{figure*}

\textbf{Foraging with four arms and in the presence of solid obstacles.}
%\medskip
%\noindent
%\textbf{Foraging with four arms in the presence of solid obstacles.}
Having established the viability of our multi-arm approach, we now consider the problem of a CyberOctopus foraging with four active arms ($I=4$). The arms, being orthogonal to each other, allow the CyberOctopus to fully traverse the substrate by crawling along the four cardinal directions, rendering the foraging problem analytically intractable. Further, this time, not only food but also solid obstacles will be distributed in 3D space (Fig.~\ref{fig:4arm-demo}a). The goal of this test is two-folds: first we wish to characterize the ability of out methods to learn to solve this foraging scenario (without obstacles), and second we wish to explore how principles of mechanical intelligence \cite{Hochner:2012, Mengaldo:2022} may be used to deal with such disturbances (obstacles), without further burdening control or training.

To this end, we incorporate a simple behavioral reflex based on traveling waves of muscle activation observed in the octopus \cite{Gutfreund:1998,Sumbre:2006}. When contact between an arm and an obstacle is sensed, two waves emanate from the point of contact in each direction. One, traveling towards the arm's tip, signals all muscles to relax, while the second, traveling towards the arm's base, signals longitudinal muscles on the contacting side to increase activation with all other muscles relaxing. Here, we treat these waves as propagating instantaneously. 
Once contact ceases, the arm returns to executing the originally prescribed muscle activations.
% are generated all muscles distal to the contact relax, while the proximal longitudinal muscle on the side in contact maximally increases its activation.
As can be seen in Fig.~\ref{fig:4arm-demo}a, this reflex, mediated by the compliant nature of the arm, allows it to slip past obstacles, thus dealing with their presence with minimal additional computational effort. On the other hand, without the reflex, the arm routinely gets stuck (Fig.~\ref{fig:4arm-demo}c).

The CyberOctopus is then initialized in the center of an arena with 40 food items randomly distributed in 3D space. Training proceeds \textit{without} obstacles using the same process of the two-arm case with the learning-based approach successfully learning to forage.
In contrast to the two-arm case, here the CyberOctopus' four active arms enable forwards/backwards and left/right crawling, allowing the collection of previously missed food, at a later stage. 
This additional freedom increases the planning required to efficiently move throughout the area, not only making an analytical Q-policy impossible to define, but also substantially impairing the ability of the greedy policy to collect food.  The greedy policy, which is here extended to deal with the case of no targets being in the workspace (see SI for details), is only able to collect 38\% of the food in the arena when engaging four arms, a notable decrease from the 54\% of food collected with two arms. Similarly to the end-to-end approach with two arms, we forgo a comparison with the random policy here, due to its substantially inferior performance.
In contrast, the PPO learning-based approach is able to successfully exploit this additional freedom (four cardinal directions of motion) to improve food collection, fetching on average 88\% of the food compared to 66\% when only two arms are active (Fig.~\ref{fig:4arm-demo}c).
The learning-based policy (trained \textit{without} obstacles) is then deployed in an environment littered with unmovable obstacles leading to arms becoming stuck, substantially impairing foraging behavior, which now only achieves 21\% food collection.
With the sensory reflex enabled, however, the CyberOctopus successfully recovers. Figure \ref{fig:4arm-demo}c shows the CyberOctopus utilizing this reflexive behavior to reach previously obstructed food (SI Video 3), resulting in 67\% of food collected. 

%%%%%%%%%%%%%%%%%%%%%%%%%%%%%%%%%%%%%%%%%%%%%%%%%%%%%%%%%%%%%%%%%%%%%%%%%%%%%%%%%%%%%%%%%%%%%%%
%%%%%%%%%%%%%%%%%%%%%%%%%%%%%%%%%%%%%%%%%%%%%%%%%%%%%%%%%%%%%%%%%%%%%%%%%%%%%%%%%%%%%%%%%%%%%%%

\section{Conclusion}
Recognizing the need for improved control methods in multi-arm soft robots, we propose a hierarchical framework inspired by the organization of the octopus neurophysiology and demonstrate it in a CyberOctopus foraging for food.
By decomposing control into high-level decision-making, low-level motor activation, and reflexive modulation via local sensory feedback and mechanical compliance, we show significant improvements relative to end-to-end approaches.
Performance is enabled via a mixed-modes approach, whereby complementary control schemes can be swapped out at any level of the hierarchy.
Here, we combine model-free reinforcement learning, for high-level decision-making, and model-based energy shaping control, for low-level muscle recruitment and activation.
To enable compatibility in terms of computational costs, we developed a novel, neural-network energy shaping (NN-ES) controller that accurately executes arm motor programs, such as reaching for food or crawling, while exhibiting time-to-solutions more than 200x faster than previous attempts \cite{Chang:2021}. 
Our hierarchical approach is successfully deployed in increasingly challenging foraging scenarios, entailing two- and three-dimensional settings, solid obstacles, and multiple arms. 

Overall, this work presents a framework to explore the control of multiple, compliant and distributed arms, in both engineering and biological settings, with the latter providing insights and hypotheses for computational corroboration. 
We have begun to take initial steps in this regard, testing how principles of mechanical intelligence and muscle waves of activation \cite{Gutfreund:1998,Sumbre:2006} may be couched into local reflexive schemes for accommodating solid obstacles. Future work will build on these foundations, exploring how distributed control approaches might enable more biologically plausible manners \cite{Hochner:2012} or how principles of mechanical intelligence may be further extended.

\medskip
\noindent\textbf{Data Access} \par \noindent
The open-source Python implementation of Elastica is available at \url{www.github.com/GazzolaLab/PyElastica}. All codes used in this paper are available at \url{https://github.com/chshih2/Real-time-control-of-an-octopus-arm-NNES} and \url{https://github.com/chshih2/Multi-arm-coordination-for-foraging}. 

% \noindent \textbf{Author Contributions:} All designed the research, C-H.S., S.H.K., N.N., U.H., H-S.C. performed the research, All wrote the paper.
% \noindent \textbf{Competing Interests:} The authors have no competing interests to declare.

\medskip
\noindent\textbf{Supporting Information} \par \noindent
Supporting information is available from the authors upon request.

% Acknowledgements
\medskip
\noindent\textbf{Acknowledgements} \par \noindent
We gratefully acknowledge Tigran Norekian and Ekaterina D. Gribkova who performed the histology of the octopus arm in Figure 2a. This study was jointly funded by ONR MURI N00014-19-1-2373 (M.G., P.G.M.), ONR N00014-22-1-2569 (M.G.), NSF EFRI C3 SoRo \#1830881 (M.G.), and  with computational support provided by the Bridges2 supercomputer at the Pittsburgh Supercomputing Center through allocation TG-MCB190004 from the Extreme Science and Engineering Discovery Environment (XSEDE; NSF grant ACI-1548562).

% References
\medskip

% \normalsize
% \linespacing

%%%%%%%%%% Insert bibliography here %%%%%%%%%%%%%%
\bibliographystyle{MSP}
% \printbibliography
\bibliography{references/Bib_Cathy.bib,references/references.bib,references/DaBib.bib}

\end{document}